\documentclass[sigconf]{acmart}

\usepackage{balance}

\def\BibTeX{{\rm B\kern-.05em{\sc i\kern-.025em b}\kern-.08emT\kern-.1667em\lower.7ex\hbox{E}\kern-.125emX}}
    
\copyrightyear{2019}
\acmYear{2019}
\setcopyright{acmcopyright}
\acmConference[MM '19] {Proceedings of the 27th ACM International Conference on Multimedia}{October 21--25, 2019}{Nice, France}
\acmBooktitle{Proceedings of the 27th ACM International Conference on Multimedia (MM'19), Oct. 21--25, 2019, Nice, France}
\acmPrice{15.00}
\acmDOI{10.1145/3343031.3350960}
\acmISBN{978-1-4503-6889-6/19/10}

\usepackage{amsmath, bm}
\usepackage{booktabs}
\begin{document}

\fancyhead{}

\title[L2G Auto-encodder]{L2G Auto-encoder: Understanding Point Clouds by Local-to-Global Reconstruction with Hierarchical Self-Attention}

\author{Xinhai Liu}
\affiliation{
\institution{School of Software, Tsinghua University \& Beijing National Research Center for Information Science and Technology (BNRist)}
\city{Beijing}
\country{China}}
\email{lxh17@mails.tsinghua.edu.cn}

\author{Zhizhong Han}
\affiliation{
\institution{Department of Computer Science, University of Maryland}
\city{College Park}
\country{USA}
}
\email{h312h@umd.edu}

\author{Xin Wen}
\affiliation{
\institution{School of Software, Tsinghua University \& Beijing National Research Center for Information Science and Technology (BNRist)}
\city{Beijing}
\country{China}
}
\email{x-wen16@mails.tsinghua.edu.cn}

\author{Yu-Shen Liu}
\authornote{Corresponding author. This work was supported by National Key R\&D Program of China (2018YFB0505400) and NSF (award 1813583).}
\affiliation{
\institution{School of Software, Tsinghua University \& Beijing National Research Center for Information Science and Technology (BNRist)}
\city{Beijing}
\country{China}
}
\email{liuyushen@tsinghua.edu.cn}

\author{Matthias Zwicker}
\affiliation{
\institution{Department of Computer Science, University of Maryland}
\city{College Park}
\country{USA}
}
\email{zwicker@cs.umd.edu}

%
\renewcommand{\shortauthors}{Liu and Han, et al.}

%
\begin{abstract}
  Auto-encoder is an important architecture to understand point clouds in an encoding and decoding procedure of self reconstruction. 
  Current auto-encoder mainly focuses on the learning of global structure by global shape reconstruction, while ignoring the learning of local structures. 
  To resolve this issue, we propose Local-to-Global auto-encoder (L2G-AE) to simultaneously learn the local and global structure of point clouds by local to global reconstruction. 
  Specifically, L2G-AE employs an encoder to encode the geometry information of multiple scales in a local region at the same time. 
  In addition, we introduce a novel hierarchical self-attention mechanism to highlight the important points, scales and regions at different levels in the information aggregation of the encoder. 
  Simultaneously, L2G-AE employs a recurrent neural network (RNN) as decoder to reconstruct a sequence of scales in a local region, based on which the global point cloud is incrementally reconstructed. 
  Our outperforming results in shape classification, retrieval and upsampling show that L2G-AE can understand point clouds better than state-of-the-art methods.
\end{abstract}

%
%

\ccsdesc[500]{Computing methodologies~Computer vision}
\ccsdesc[500]{Computing methodologies~Shape representations}
\ccsdesc[300]{Information systems~Information retrieval}

%
\keywords{point clouds; auto-encoder; unsupervised learning; hierarchical attention; interpolation layer; recurrent neural network}

\maketitle

\begin{figure*}[htp]
  \centering
  \includegraphics[height=5cm,width=16cm]{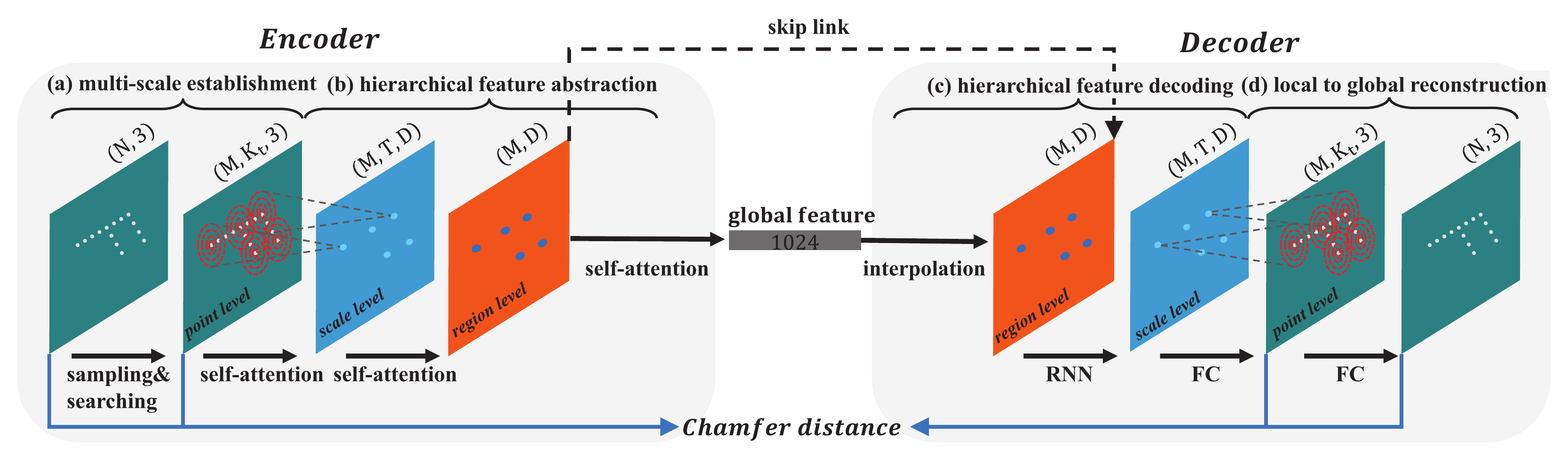}
  \caption{Illustration of our local to global auto-encoder architecture. In the encoder, multi-scale areas is established in each local region around the sampled centroids in (a). And a hierarchical feature abstraction is employed to abstract the global feature of point clouds with self-attention in (b). The learned global feature is applied to shape classfication and retrieval applications.  In the decoder, local areas and the global point cloud are reconstructed by hierarchical feature decoding with the interpolation layer, the RNN layer and the FC layer in (c)(d).}
  \label{fig:architecture}
\end{figure*}
\section{Introduction}
In recent years, point clouds have attracted increasing attention due to the popularity of various depth sensors in different applications. Not only the traditional methods, deep neural networks have also been applied to point cloud analysis and understanding.
However, it remains a challenge to directly learn from point clouds.
Different from 2D images, point cloud is an irregular 3D data  which makes it difficult to directly use traditional deep learning framework, e.g., traditional convolution neural network (CNN).
The traditional CNN usually requires some fixed spatial distribution around each pixel so as to facilitate the convolution.
One way to alleviate the problem is to voxelize a point cloud into voxels and then apply 3D Cov-Nets.
However, because of the sparsity of point clouds, it leads to resolution-loss and explosive computation complexity, which sacrifices the representation accuracy.

To address above challenges, PointNet  \cite{qi2016pointnet} has been proposed to directly learn shape representations from raw point sets. Along with the availability of directly learning from point clouds by deep learning models, auto-encoder (AE) has become an vital architecture of the involved neural networks. Current AE focuses on the learning of the global structure of point clouds in the encoding and decoding procedure. However, current AE structure is still limited by learning the local structure of point clouds, which tends to be an important piece of information for point cloud understanding.

To simultaneously learn global and local structure of point clouds, we propose a novel auto-encoder called Local-to-Global auto-encoder (L2G-AE). Different from traditional auto-encoder, L2G-AE leverages a local region reconstruction to learn the local structure of a point cloud, based on which the global shape is incrementally reconstructed for the learning of the global structure. 
Specifically, the encoder of L2G-AE can hierarchically encode the information at point, scale and region levels, where a novel hierarchical self-attention is introduced to highlight the important elements in each level.
The encoder further aggregates all the information extracted from the point cloud into a global feature. 
In addition, L2G-AE employs a RNN-based decoder to decode the learned global feature into a sequence of scales in each local region.
And based on scale features, the global point cloud is incrementally reconstructed. 
L2G-AE leverages this local to global reconstruction to facilitate the point cloud understanding, which finally enables local and global reconstruction losses to train L2G-AE.

Our key contributions are summarized as follows.
\begin{itemize}
    \item We propose L2G-AE to enable the learning of global and local structures of point clouds in an auto-encoder architecture, where the local structure is very important in learning highly discriminative representations of point clouds.
    \item We propose hierarchical self-attention to highlight important elements in point, scale and region levels by learning the correlations among the elements in the same level.
    \item We introduce RNN as decoding layer in an auto-encoder architecture to employ more detailed self supervision, where the RNN takes the advantage of the ordered multi-scale areas in each local region.
\end{itemize}

\section{Related Work}
Point clouds is a fundamental type of 3D data format which is very close to the raw data of various 3D sensors.
Recently, applications of learning directly on point clouds  have received extensive attention, including shape completion \cite{Stutz2018CVPR}, autonomous driving \cite{qi2017frustum}, 3D object detection \cite{simon2018complex,Yang2018pixor,zhou2017voxelnet}, recognition and classification \cite{qi2016pointnet,NIPS2017_7095,golovinskiy2009shape,li2018pointcnn,wang2018dynamic,xu2018spidercnn,shen2018mining,xie2018attentional,li2018so,you2018pvnet}, scene labeling \cite{NIPS2011_4226}, upsampling \cite{yu2018pu,2019arXiv181111286Y}, dense labeling and segmentation \cite{Wang2018pointseg} , etc.

Due to the irregular property of point cloud and the inspiring performances of 2D CNNs on large-scale image repositories such as ImageNet \cite{deng2009imagenet}, it is intuitive to rasterize point clouds into 3D voxels and then apply 3D CNNs.
Some studies \cite{qi2017frustum,zhou2017voxelnet,HanCyber17a} represent each voxel with a binary value which indicates the occupation of this location in space.
The main problem of voxel-based methods is the fast growth of neural network size and computation complexity with the increasing of spatial resolution.
To alleviate this problem, some improvements \cite{li2016fpnn} have been proposed to explore the data sparsity of point clouds.
However, when dealing with point clouds with huge number of points, the complexity of the neural network is still unacceptable.

Recently, deep neural networks work quite effectively on the raw 3D point clouds.
Different from learning from readered views \cite{han2018seqviews2seqlabels,han20193d2seqviews,han20192seq2seq,han2019view,han2019parts4feature,han20193dviewgraph} 2D meshes \cite{Zhizhong2016} or 3D voxels \cite{Zhizhong2016b,han2017boscc,han2018deep}, PointNet \cite{qi2016pointnet} is the pioneer study which directly learns the representation for point clouds by computing features for each point individually and aggregating these features with max-pool operation.
To capture the contextual information of local patterns inside point clouds, PointNet++ \cite{NIPS2017_7095} uses sampling and grouping operations to extract features from point clusters hierarchically.
Similarly, several recent studies \cite{riegler2017octnet,klokov2017escape} explores indexing structures, which divides the input point cloud into leaves, and then aggregates node features from leaves to the root.
Inspired by the convolution operation, recent methods \cite{li2018pointcnn,wang2018dynamic,xu2018spidercnn} investigate well-designed CNN-like operations to aggregate points in local regions by building local connections with k-neareat-neighbors (kNN).

Capturing the context information inside local regions is very important for the discriminative ability of the learned point cloud representations.
KC-Net \cite{shen2018mining} employs a kernel correlation layer and a graph pooling layer to capture the local patterns of point clouds.
ShapeContextNet \cite{xie2018attentional} extends 2D Shape Context \cite{belongie2001shape} to the 3D, which divides a local region into small bins and aggregates the bin features.
Point2Seqeuce \cite{liu2019point2sequence} employs an attention-based sequence to sequence architecture to encode the multi-scale area features inside local regions.

In order to alleviate the dependence on the labeled data, some studies have performed unsupervised learning for point clouds.
FoldingNet \cite{yang2018foldingnet} proposes a folding operation to deform a canonical 2D grid onto the suface of a point cloud.
3D-PointCapsNet \cite{zhao2019pcn} employs a dynamic routing scheme in the reconstruction of input point clouds.
However, it is difficult for these methods to capture the local patterns of point clouds.
Similar to FoldingNet, PPF-FoldNet \cite{Deng2018ppffoldnet} also learns local descriptors on point cloud with a folding operation.
LGAN \cite{Achlioptas2017latent} proposes an auto-encoder based on PointNet and extends the decoder module to the point cloud generation application with GAN.
In this work, we propose a novel auto-encoder architecture to learn representations for point clouds.
On the encoder side, an hierarchical self-attention mechanism is applied to embedding the correlation among features in each level.
And on the decoder side, an interpolation layer and a RNN decoding layer are engaged to reconstruct multi-scale areas inside local regions.
After building local areas, the global point cloud is generated by a fully-connected (FC) layer which acts as a down sampling function.

\section{Method}
Now we introduce the L2G-AE in detail, where the structure is illustrated in Figure \ref{fig:architecture}.
The input of the encoder is an unordered point set $\bm{P} = \{p_1, p_2, \cdots, p_N \}$ with $N$ ($N=1024$) points.
Each point in the point set is composed of a 3D coordinate $(x,y,z)$.
L2G-AE first establishes multi-scale areas $\bm{A}_t$ $(t \in [1,T])$ in each local region around the sampled points.
Then, a hierarchical feature abstraction is enforced to obtain the global features of input point clouds with self-attentions.
In the decoder, we simultaneously reconstruct local scale areas and global point clouds by hierarchical feature decoding.
The output of L2G-AE is the reconstructed local areas $\bm{A}^{'}_t$ and the reconstructed $\bm{P}^{'}$ with same number of points to $\bm{P}$.
\begin{figure}
  \centering
  \includegraphics[height=3.5cm,width=5cm]{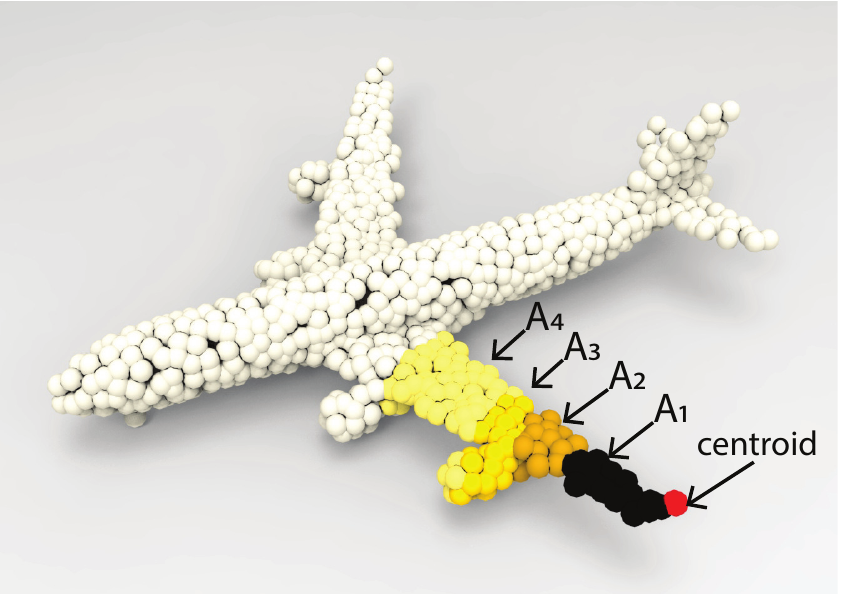}
  \caption{A multi-scale example inside a local region of an airplane point cloud, where there are four scales areas $[\bm{A}_1, \bm{A}_2, \bm{A}_3, \bm{A}_4]$ with different colors around the centroid point (red).}
  \label{fig:multiscale}
\end{figure}

\subsection{Multi-scale Establishment}
To capture fine-grained local patterns of point clouds, we first establish multi-scale areas in each local region, which is similar to PointNet++ \cite{NIPS2017_7095} and Point2Sequence \cite{liu2019point2sequence}.
Firstly, a subset $\{p_{i_1}, p_{i_2},\\ \cdots, p_{i_M}\}$ of the input points is selected as the centroid of local regions by iterative farthest point sampling (FPS).
The latest point $p_{i_j}$ is always the farthest one from the rest points $\{p_{i_1}, p_{i_2}, \cdots, p_{i_{j-1}}\}$.
Compared to other sampling method, such as random sampling, FPS can achieve a better coverage of the entire point cloud with the given same number of centroids.  
As shown in Figure \ref{fig:multiscale}, around each sampled centroid, $T$ different scale  local areas are established continuously by kNN searching with $\{K_1, K_2, \cdots, K_T\}$  nearest points, respectively.  
An alternative searching method is ball query \cite{NIPS2017_7095} which selects all points with a radius around the centroid.
However, it is difficult for ball query to ensure the information inside local regions, which is sensitive to the sparsity of the input point clouds.

\begin{figure}[htp]
  \centering
  \includegraphics[height=4cm,width=8cm]{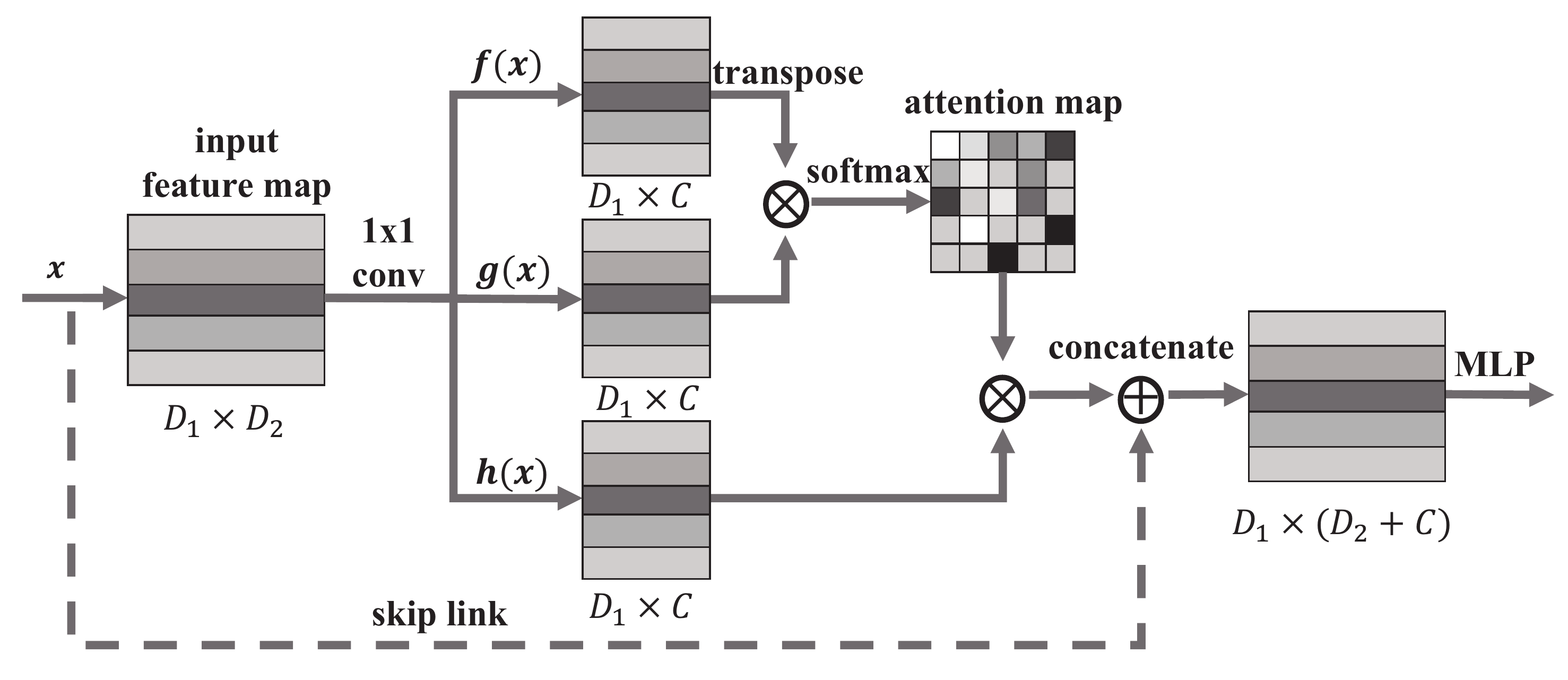}
  \caption{Self-attention module. The input of this module is a $D_1 \times D_2$ feature map and the output is another $D_1 \times (D_2 + C)$ feature map, where $C$ is a parameter.}
  \label{fig:c_e}
\end{figure}
\subsection{Hierarchical Self-attention}
In current work of learning on point clouds, Multi-Layer-Perceptron (MLP) layer is widely applied to integrate multiple features.
Traditional MLP layer first abstracts each feature into higher dimension individually and then aggregates these features by a concise max pooling operation.
However, these two simple operations can hardly encode the correlation between feature vectors in the feature space.
Inspired by the self-attention machanism in \cite{zhang2018sa}, the attention machanism is suitable for improving the traditional MLP by learning the correlation between features.
In this work, we propose a self-attention module to make up the defects of the MLP layer with an attention mechanism.
Here, self-attention refers to learn the correlation among features in the same level.

Different from the raw self-attention, we enforce a hierarchical feature extraction architecture with hierarchical self-attention in the encoder.
There are three different levels inside the encoder, including point level, scale level, and region level.
At each level, we introduce a self-attention module to learn self-attention weights by mining the correlations among the corresponding feature elements.
Consequently, three self-attention modules are designed to propagate features from the lower level to the higher level.
Supposed the input of the self-attention module is a feature map $\bm{x} \in \mathbb{R}^{D_1 \times D_2}$, where $D_1$, $D_2$ are the dimensions of the feature map.
Therefore, $D_1$, $D_2$ are equal to $K_t$, $3$ in the point level, equal to $T$, $D$ in the scale level and equal to $M$, $D$ in the region level, respectively.

As depicted in Figure \ref{fig:c_e}, the feature map  $\bm{x}$ is first transformed into two feature spaces $\bm{f}$ and $\bm{g}$ to calculate the attention below, where $\bm{f}(\bm{x}) = \bm{W_f x}$, $\bm{g}(\bm{x}) = \bm{W_g x}$,
\begin{equation}
    \beta_{j,i} = \frac{exp(s_{ij})}{\sum_{i=1}^{D_1} exp(s_{ij})}, \text{where } s_{ij} = \bm{f}(\bm{x_i})^T\bm{g}(\bm{x_j}),
\end{equation}
and $\beta_{j,i}$ evaluates the attention degree which the model pays to the $i^{th}$ location when synthesizing the $j^{th}$ feature vector.
Then the attention result is $\bm{r} = (\bm{r_1}, \bm{r_2}, \cdots, \bm{r_j}, \cdots, \bm{r_{D_1}}) \in \mathbb{R}^{D_1 \times D_2}$, where
\begin{equation}
    \bm{r_j} = \sum_{i=1}^{D_1} \beta_{j,i}\bm{h}(\bm{x_i}), \text{where }\bm{h}(\bm{x_i}) = \bm{W_h x_i}.
\end{equation}
In above formulation, $\bm{W_f},\bm{W_g},\bm{W_h} \in \mathbb{R}^{D_2 \times C}$ are learned weight matrices, which are implemented as $1\times1$ convolutions.
We use $C = M / 8 $ in the experiments.

In addition, inspired by the skip link operation in ResNet\cite{he2016deep} and DenseNet \cite{huang2017densely}, we further concatenate the result of the attention mechanism with the input feature matrix.
Therefore, the final output of the self-attention module is given by
\begin{equation}
    \bm{o_i} = \bm{x_i} \oplus \bm{r_i},
\end{equation}
where $\oplus$ is the concatenation operation.
This allows the network to rely on the cues among the feature vectors.

To aggregate the features with correlation information, a MLP layer  and a max pooling operation are employed to integrate the multiple features.
In particular, the first self-attention module aggregates the points in a scale to a D-dimensional feature vector.
The second one encodes the multi-scale features in a region into a D-dimensional feature.
The final one integrates features of all local regions on a point cloud into a 1024-dimensional global feature.
Therefore, the encoder hierarchically abstracts point features from the levels of point, scale and region to a global representation of the input point cloud.

\subsection{Interpolation Layer}
The target of the decoder is to generate the points of the local areas and entire points.
Previous approaches \cite{Achlioptas2017latent,yang2018foldingnet,Deng2018ppffoldnet} usually use simple fully-connected (FC) layers or MLP layers to build the decoder.
However, the expressive ability of the decoder is largely limited without considering the relationship among features.
In this work, we propose a progressive decoding way which can be regarded as a reverse process of the encoding.
The first step is to generate local region features from the global feature.
To propagate the global feature $\bm{g}$ to region features, a simple interpolation operation is first engaged in the decoder. The local region feature $\bm{l}_i$ is calculated by
\begin{equation}
\bm{l}_i = \frac{c}{(p_i-p_0)^2}\bm{g}, i \in [1,M],
\end{equation}
where $c$ ($c={10}^{-10}$) is a constant.
Here, $p_0 = (0,0,0)$ is the centroid of the input point cloud after the normalization processing.
And $p_i$ is the centroid point of the corresponding local region.
By the simple interpolation operation, the spatial distribution information of local region can be integrated to facilitate the feature decoding.
The interpolated local region features are then concatenated with skip linked local region features from the encoder.
The concatenated features are passed through another MLP layer into a $M \times D$ feature matrix.
\begin{figure}[htp]
    \centering
    \includegraphics[height=2cm,width=8cm]{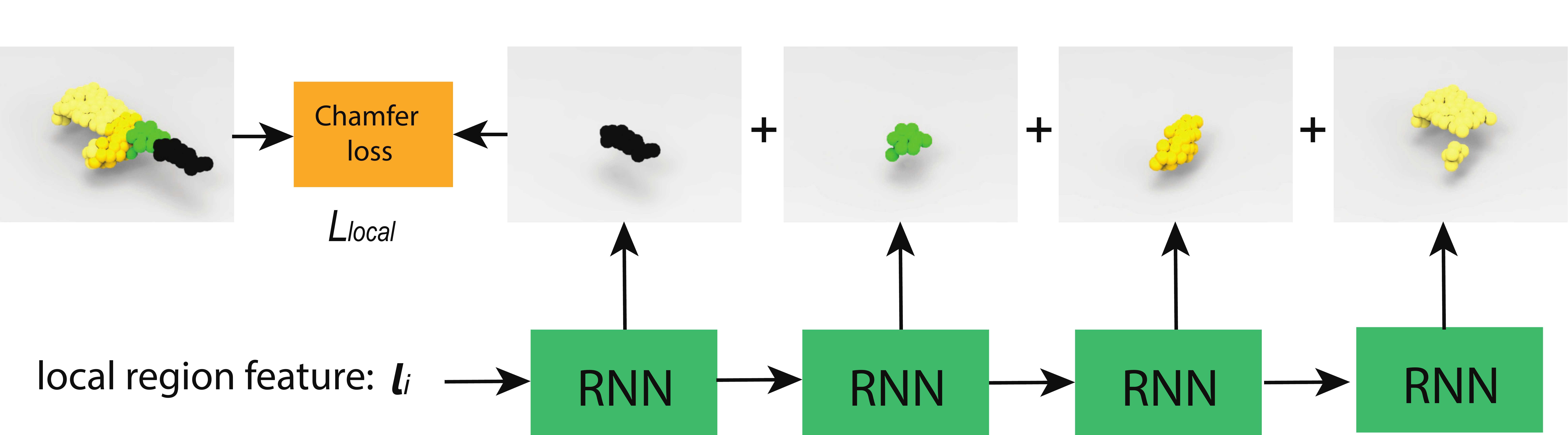}
    \caption{The decoding process of the RNN layer.}
    \label{fig:rnn}
\end{figure}
\subsection{RNN Layer}
Given the feature of local regions, we want to decode the scale level features.
Due to the multi-scale setting, the features of different scales in a local region can be regarded as a feature sequence with length $T$.
As we all know that recurrent neural network \cite{hochreiter1997long} has shown excellent performances in processing sequential data.
Thus, a RNN decoding layer is employed to generate the multi-scale area features.
The decoding process is shown in Figure \ref{fig:rnn}.
We first replicate the local region feature $\bm{l}_i$ for $T$ times, and the replicated local region features are feed into the RNN layer by
\begin{equation}
    \bm{h}_t = f(\bm{h}_{t-1}, \bm{l}_i^t), t \in [1,T],
\end{equation}
where $f$ is a non-linear activation function and $t$ is the index of RNN step.
Therefore, the predicted $t^{th}$ area feature $\bm{a}_t$ can be calculated by
\begin{equation}
   \bm{a}_t = \bm{W}_\theta \bm{h}_t.
\end{equation}
Here, $\bm{W}_d$ is a learnable weight matrix.
To generate the points inside each local area, several FC layers are adopted to reconstruct the points.
The local area $\bm{A}^{'}_t$ is reconstructed by
\begin{equation}
    \bm{A}^{'}_t = \bm{W}_{\theta_t}\bm{a}_t + b_{\theta_t},
\end{equation}
where $\bm{W}_{\theta_t}$, $b_{\theta_t}$ are weights of the FC layer.
Based on the reconstructed local areas, another FC layer is applied to incrementally reconstruct the entire point cloud.
All reconstructed areas are concatenated and then passed through the FC layer by
\begin{equation}
    \bm{P} = \bm{W} [\bm{A}^{'}_1 \oplus \bm{A}^{'}_2 \oplus \cdots \oplus \bm{A}^{'}_T] + b.
\end{equation}
Here, $\oplus$ represents the concatenation operation.
\subsection{Loss Function}
We propose a new loss function to train the network in an end-to-end fashion.
There are two parts in the loss function, local scale reconstruction and global point cloud reconstruction, respectively.
As mentioned earlier, we should encourage accurate reconstruction of local areas and the global point cloud at the same time.
Suppose $\bm{A}_t$ is the $t^{th}$ scale area in the multi-scale establishment subsection, then, the local reconstruction error for $\bm{A}^{'}_t$ is measured by the well-known Chamfer distance,
\begin{equation}
\begin{split}
L_{local} = d_{CH}(\bm{A}_t, \bm{A}^{'}_t) = \sum_{t=1}^T  (\frac{1}{|\bm{A}_t|}\sum_{p_i \in \bm{A}_t} \min_{p^{'}_i \in \bm{A}^{'}_t} \lVert p_i - p^{'}_i \rVert_2  \\ +  \frac{1}{|\bm{A}^{'}_t|}\sum_{p^{'}_i \in \bm{A}^{'}_t} \min_{p_i \in \bm{A}_t} \lVert p_i - p^{'}_i \rVert_2),
\end{split}
\end{equation}
Similarly, let the input point set be $\bm{P}$ and the reconstructed point set be $\bm{P}^{'}$. The global reconstruction error can be denoted by
\begin{equation}
\begin{split}
     L_{global} = d_{CH}(\bm{P}, \bm{P}^{'}) = \frac{1}{|\bm{P}|}\sum_{p_i \in \bm{P}} \min_{p^{'}_i \in \bm{P}^{'}} \lVert p_i - p^{'}_i \rVert_2 \\
     + \frac{1}{|\bm{P}^{'}|}\sum_{p^{'}_i \in \bm{P}^{'}} \min_{p_i \in \bm{P}} \lVert p_i - p^{'}_i \rVert_2.
\end{split}
\end{equation}
Altogether,  the network is trained end-to-end by minimizing the following joint loss function
\begin{equation}
    L = L_{local} + \gamma L_{global},
\end{equation}
where $\gamma$ ($\gamma=1$) is the proportion of two part errors.
\section{Experiments}
In this section, we first investigate how some key parameters affect the performance of L2G-AE in the shape classification task on ModelNet10 \cite{wu20153d}.
Then, an ablation study is done to show the effectiveness of each module in  L2G-AE.
Finally, we further evaluate the performances of L2G-AE in multiple applications including 3D shape classification, 3D shape retrieval and point cloud upsampling.
\subsection{Network Configuration}
In L2G-AE, we first sample $M = 256$ points as the centroids of local regions by FPS.
Then, around each centroid, a kNN searching algorithm selects $T=4$ scale areas with $[K_1=16, K_2=32, K_3=64, K_4=128]$ points inside each area.
In the multi-level feature propagation process, we initialize the feature  dimension $C = M/8 = 32$ and $D = 256$.
The encoder learns a 1024-dimension global feature for the input point cloud through hierarchical feature extraction.
Similarly, the decoder hierarchically reconstructs local scales and global point cloud.
In the RNN decoding layer, we adopt LSTM as the default RNN cell with hidden state dimension $h = D = 256$.
In the experiment, we train our network on a NVIDIA GTX 1080Ti GPU using ADAM optimizer with the initial learning rate of 0.0001 and batch size of 8.
The learning rate is decreased by 0.3 for every 20 epochs.
\subsection{Parameters}
All experiments on parameter comparison are evaluated under ModelNet10. ModelNet10 contains 4899 CAD models from 10 categories and is split into 3991 for training and 908 for testing.
For each model, we adopt 1024 points which are uniformly sampled from mesh faces and are normalized into a unit ball before being fed into the network.
During the training process, the loss function keeps decreasing and stabilizes around the 180th epoch.
To acquire the accuracies on ModelNet10, we train a linear SVM from the global features obtained by the auto-encoder.
Specifically, the OneVsRest strategy is adopted with the linearSVM function as the kernel.

We first explore the number of sampled points $M$ which determines the distribution of local regions inside point clouds.
In the experiment, we keep the network settings as depicted in the network configuration and vary the number of sampled points $M$ from 128 to 320.
\begin{table}[htp]
\centering
\caption{The effects of the number of sampled points $M$ under ModelNet10.}
\label{table:sampled_points}
\begin{tabular}{cccccc}\hline
$M$ &128 &192 &256 &320 \\ \hline
Acc (\%)&93.83 &94.38 &\textbf{95.37} &93.94\\ \hline
\end{tabular}
\end{table}
The results are shown in Table \ref{table:sampled_points}, where the instance accuracies on the benchmark of ModelNet10 have a tendency to rise first and then fall.
This comparison implies that L2G-AE can effectively extract the contextual information in point clouds by multi-level feature propagation and $M=256$ is an optimal choice which can well cover input point clouds without excessive redundant.
\begin{figure}
    \centering
    \includegraphics[height=1.5cm,width=8cm]{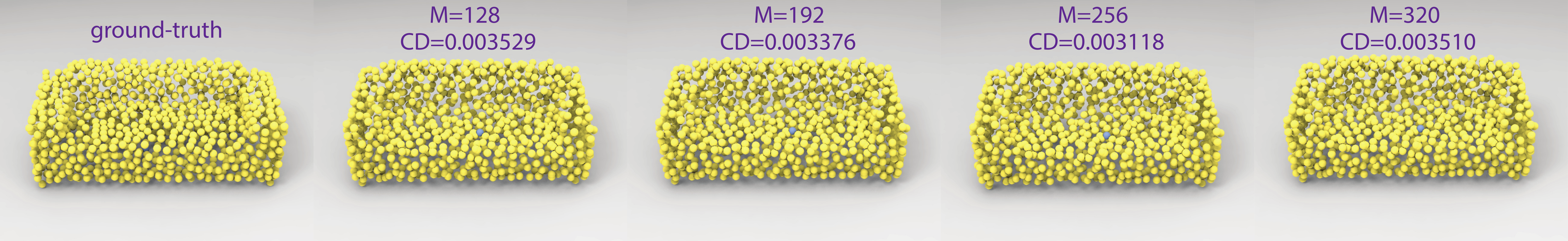}
    \caption{The reconstructed results with different sampled points, where the CD represents the Chamfer distance between ground-truth and the reconstructed point cloud.}
    \label{fig:sample_re}
\end{figure}
To learn the reconstructed results intuitively, Figure \ref{fig:sample_re} shows the reconstructed point clouds with different sampled points.
According to Chamfer distances, L2G-AE can also reconstruct the input point cloud with the varying of sampled points.

With keeping the sampled points $M = 384$,  we investigate the key parameter dimension $C$ inside the self-attention modules.
To unify the parameter in self-attention module, we keep the same dimension $C$ in different semantic levels.
We change the default $C = 32$ to 16 and 64, respectively.
In Table \ref{table:c}, L2G-AE achieves the best performance when the feature dimension $C$ is 32.
\begin{table}[htp]
\centering
\caption{The effects of the feature dimension $C$ of the self-attention module under ModelNet10.}
\label{table:c}
\begin{tabular}{ccccc}\hline
$M$ &16 &32 &64 \\ \hline
Acc (\%) &93.94 &\textbf{95.37} &94.16 \\ \hline
\end{tabular}
\end{table}
Finally, we show the effects of feature dimension of local areas $D$ and the global feature $D_{global}$.
The dimension is varied as shown in Table \ref{table:d} and Table \ref{table:dglobal}.
Neither the biggest nor the smallest, L2G-AE gets better performances when $D$, $D_{global}$ are set to 256 and 1024 respectively.
There is a trade-off between the network complexity and the expressive ability of our L2G-AE.
\begin{table}[htp]
\centering
\caption{The effects of the local feature dimension $D$ on ModelNet10.}
\label{table:d}
\begin{tabular}{ccccc}\hline
$D$ &128 &256 &512 \\ \hline
Acc (\%) &93.72 &\textbf{95.37} &93.28 \\ \hline
\end{tabular}
\end{table}
\begin{table}[htp]
\centering
\caption{The effects of the global feature dimension $D_{global}$ under ModelNet10.}
\label{table:dglobal}
\begin{tabular}{ccccc}\hline
$D_{global}$ &512 &1024 &2048 \\ \hline
Acc (\%) &94.16 &\textbf{95.37} &93.94 \\ \hline
\end{tabular}
\end{table}
\subsection{Ablation Study}
To quantitatively evaluate the effect of the self-attention module, we show the performances of L2G-AE under four settings: with point level self-attention module only (PL), with area level self-attention module only (AL), with region level self-attention module only (RL), remove all self-attention modules (NSA) and with all self-attention modules (ASA). As shown in Table \ref{table:rsa}, the self-attention module is effective in learning highly discriminative representations of point clouds by capturing the correlation among feature vectors.
The results with only one self-attention module outperform the results without any self-attention module.
And we achieve the best performance when three self-attention modules work together.
The performance of self-attentions is affected by the discriminative ability of features. 
At the area level, the features of areas in the same region are similar, since there are only four areas, which makes the self-attention at area level contribute the least among all three self-attentions.  
In contrast, at the point level and the region level, the features of points or regions change a lot, so these self-attentions contribute more. 
From our observation, the results of PL and RL are coincidentally equal in the experiments. 
\begin{table}[htp]
\centering
\caption{The effects of the self-attention module on ModelNet10.}
\label{table:rsa}
\begin{tabular}{cccccc}\hline
Metric&PL&AL&RL&NSA&ASA\\ \hline
Acc (\%)&94.16& 94.05&94.16&93.72&\textbf{95.37}\\ \hline
\end{tabular}
\end{table}

\begin{figure}
  \centering
  \includegraphics[width=8cm,height=2cm]{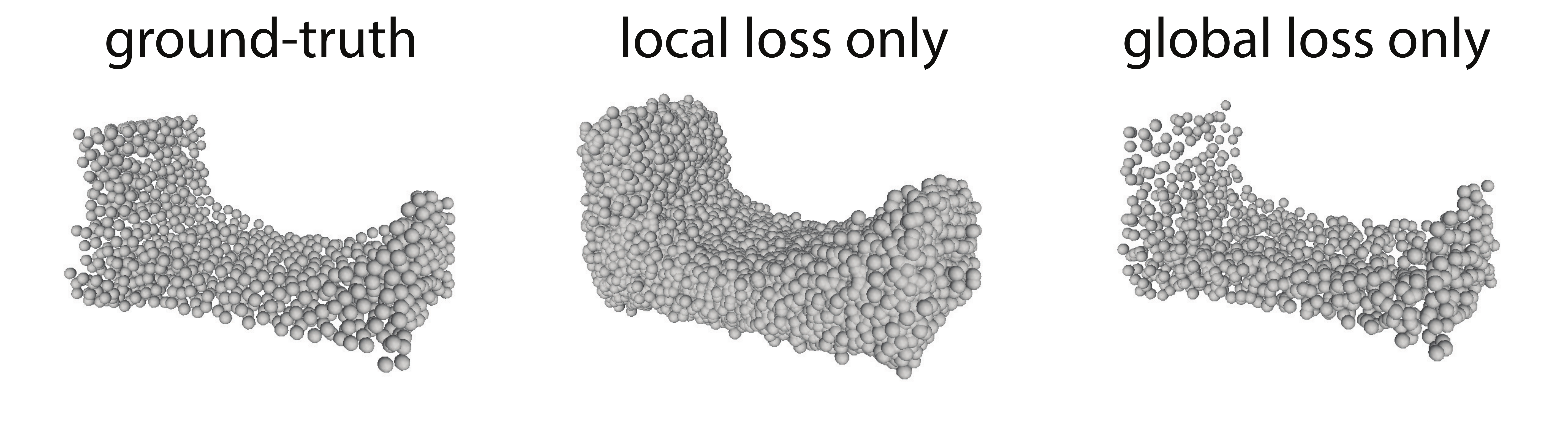}
  \caption{The reconstruction results of L2G-AE with only the local loss and only the global loss.}\label{fig:loss_compare}
\end{figure}
After exploring the self-attention module, we also discuss the contributions of the two loss functions $L_{local}$ and $L_{global}$.
In Table \ref{table:loss}, the results with local loss only (Local), global loss only (Global) and  two losses together (Local + Global) are listed.
The local loss function is very important in capturing local patterns of point clouds.
And the two loss functions together can further enhance the classification performances of our neural network.
In addition, Figure \ref{fig:loss_compare} shows the reconstruction results of our L2G-AE with only local loss and only global loss, respectively.
From the results of the reconstructed point clouds, L2G-AE can reconstruct the input point cloud with only part of the joint loss function.
In particular, the local reconstructed result in Figure \ref{fig:loss_compare} is a dense point cloud.
\begin{table}[htp]
\centering
\caption{The effects of the two loss functions $L_{local}$ and $L_{global}$ on ModelNet10.}
\label{table:loss}
\begin{tabular}{cccc}\hline
Metric&Local&Global&Local+Global\\ \hline
Acc (\%)&94.71 &92.84 &\textbf{95.37}\\ \hline
\end{tabular}
\end{table}

\begin{table}[htp]
\centering
\caption{The comparison of classification accuracy (\%) under  ModelNet10 and ModelNet40.}
\label{table:compare}
\begin{tabular}{ccccc}
\hline
Methods          &Supervised &MN40 &MN10 \\ \hline
 PointNet        &Yes  &89.20   &-       \\
 PointNet++      &Yes  &90.70   &-       \\
 ShapeContextNet &Yes  &90.00	&-       \\
 Kd-Net     	 &Yes  &91.80   &94.00   \\
 KC-Net   		 &Yes  &91.00   &94.4    \\
 PointCNN    	 &Yes  &92.20   &-       \\
 DGCNN       	 &Yes  &92.20   &-       \\
 SO-Net          &Yes  &90.90   &94.1    \\
 Point2Sequence  &Yes  &92.60   &95.30   \\ \hline
 MAP-VAE         &No   &90.15   &94.82  \\
 LGAN            &No   &85.70   &95.30   \\
 LGAN(MN40)      &No   &87.27   &92.18   \\
 FoldingNet            &No   &88.40   &94.40   \\
 FoldingNet(MN40)      &No   &84.36   &91.85   \\ 
\hline
 Our             &No   &\textbf{90.64} &\textbf{95.37} \\
\hline
\end{tabular}
\end{table}
\subsection{Classification}
In this subsection, we evaluate the performance of L2G-AE under ModelNet10 and ModelNet40 benchmarks, where
ModelNet40 contains 12, 311 CAD models which is split into 9, 843 for training and 2, 468 for testing.
Table \ref{table:compare} compares L2G-AE with state-of-the-art methods in the shape classification task on ModelNet10 and ModelNet40.
The compared methods include PointNet \cite{qi2016pointnet}, PointNet++ \cite{NIPS2017_7095}, ShapeContextNet \cite{xie2018attentional}, KD-Net \cite{klokov2017escape}, KC-Net \cite{shen2018mining}, PointCNN \cite{li2018pointcnn}, DGCNN \cite{wang2018dynamic}, SO-Net \cite{li2018so}, Point2Sequence \cite{liu2019point2sequence}, MAP-VAE \cite{Zhizhong2019mapvae}, LGAN \cite{Achlioptas2017latent} and FoldingNet \cite{yang2018foldingnet}.

L2G-AE significantly outperforms all the unsupervised competitors under ModelNet10 and ModelNet40, respectively.
In particular, L2G-AE achieves accuracy $95.37\%$ which is even higher than other methods of supervision under ModelNet10.
Although the results of LGAN \cite{Achlioptas2017latent} and FoldingNet \cite{yang2018foldingnet}  also show good performances under ModelNet10 and ModelNet40.
This is because these methods are trained under a version of ShapeNet55 that contains more than 57,000 3D shapes.
However, this version of ShapeNet55 dataset is not avaiable for public download from the official website.
Therefore, we train all these methods under ModelNet40 for the fair comparison.

\begin{table}[htp]
\centering
\caption{The comparison of retrieval in terms of under ModelNet10.}
\label{table:retrieval}
\begin{tabular}{cccc}
\hline
Methods             & LGAN    & FoldingNet  & Our    \\ \hline
Acc (\%)            &49.94      &53.42       &\textbf{67.81}    \\
\hline
\end{tabular}
\end{table}
\begin{figure}
    \centering
    \includegraphics[height=3cm,width=4cm]{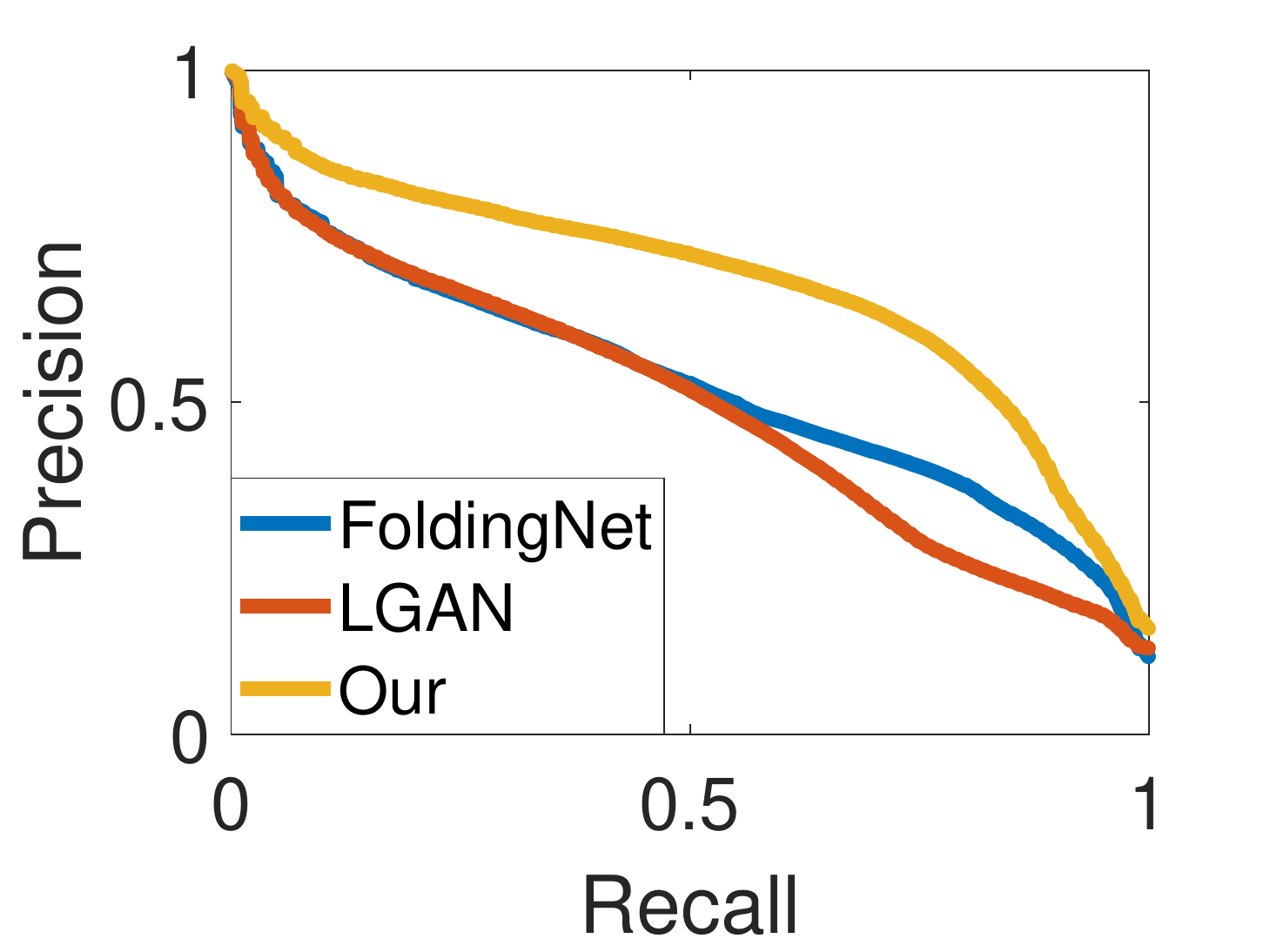}
    \caption{The comparison of PR curves for retrieval under ModelNet10.}
    \label{fig:pr}
\end{figure}
\begin{table*}[htp]
\centering
\caption{The quantitative comparison of $16 \times$ upsampling from 625 points under ModelNet10.}
\label{table:upsampling}
\begin{tabular}{cccccccccccc}
\hline
$10^{-3}$ &bathtub &bed &chair &desk &dresser &monitor &n.stand & sofa &table &toilet              \\ \hline
 PU      &\textbf{1.01} &\textbf{1.12} &\textbf{0.82} &\textbf{1.22} &1.55 &\textbf{1.19} &\textbf{1.77} &\textbf{1.13} &\textbf{0.69} &\textbf{1.39 }            \\
 EC      &1.43 &1.81 &1.80 &1.30 &1.43 &2.04 &1.88 &1.79 &1.00 &1.72            \\
 Our     &1.74 &1.46 &1.58	&2.08 &\textbf{1.40} &1.61 &1.86 &1.67 &1.86 &2.10
   \\
\hline
\end{tabular}
\end{table*}
\subsection{Retrieval}
L2G-AE is further evaluated in the shape retrieval task under ModelNet10 and compared with some other unsupervised methods of learning on point clouds.
The compared results include two state-of-the-art unsupervised methods for point clouds, i.e., LGAN \cite{Achlioptas2017latent} and FoldingNet \cite{yang2018foldingnet}.
The target of shape retrieval is to obtain the relevant information of a inquiry from a collection.
In these experiments, the 3D shapes in the test set are used as quires to retrieve the rest shapes in the same set, and mean Average Precision (mAP) is used as a metric.

As shown in Table \ref{table:retrieval}, our results outperform all the compared results under ModelNet10.
It shows that L2G-AE can be effect in improving the performance of unsupervised shape retrieval on point clouds.
Their PR curves under ModelNet10 are also compared in Figure \ref{fig:pr} which intuitively shows the performances of these three methods.

\subsection{Unsupervised  Upsampling for Point Clouds}
Benefit from the design of local to global reconstruction,  it is competent for our L2G-AE to be applied in the unsupervised point cloud upsampling application.
In the local reconstruction, a dense point cloud is obtained by reconstructing multiple local scales with overlapping.
Therefore, it is convenient to produce the upsampling results by downsampling from the dense local reconstructed results using some unsupervised methods, such as random sampling or farthest point sampling.
As far as we know, L2G-AE is the first method which performs point cloud upsampling with deep neural networks in an unsupervised manner.
To evaluate the performance of L2G-AE, We compare our method on relatively sparse (625 points) inputs with state-of-the-art supervised point cloud upsampling methods, including PU-Net \cite{yu2018pu} and EC-Net \cite{yu2018ecnet}.
The target of upsampling is to generate a dense point clouds with 10000 points.
For PU-Net and EC-Net, the $16 \times$ results (10000 points) are obtained from inputs (625 points) in a supervised manner.
Differently, L2G-AE first obtains the local reconstruction results and then downsamples them to 10000 points.

\begin{figure}
  \centering
  \includegraphics[width=8cm, height=6cm]{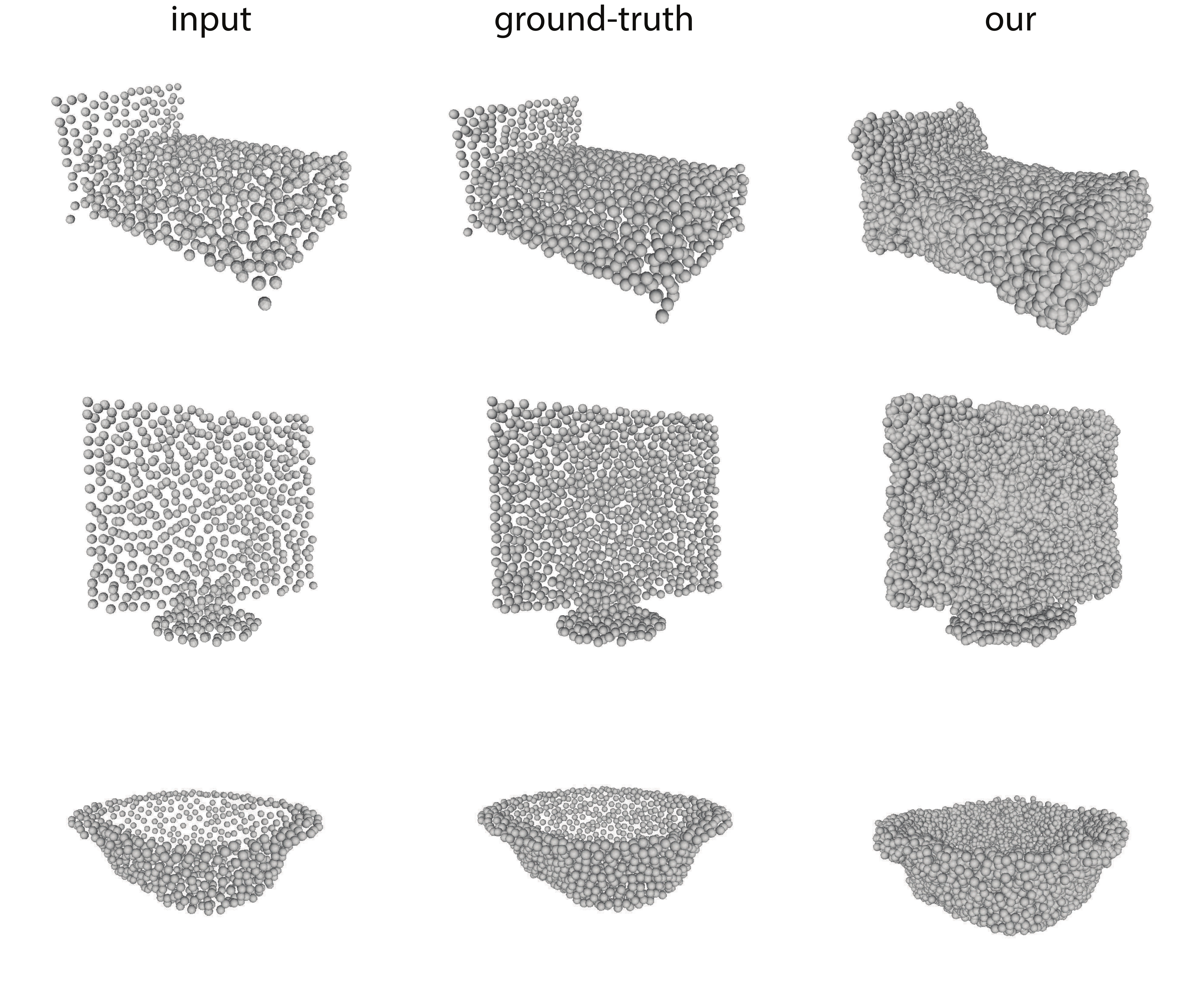}
  \caption{Some upsampled results of L2G-AE.}\label{fig:upsampled}
\end{figure}
As shown in Table \ref{table:upsampling}, mean Chamfer Distance (mCD) is used as a metric for quantitative comparison with PU-Net (PU) and EC-Net (EC) under ModelNet10.
Although the results of PU-Net and EC-Net are better than "Our" in some classes under ModelNet10, the most likely reason is that the ground-truth is not visible to L2G-AE in the training.
In addition, the input point cloud with 625 points contains very limited information.
Figure \ref{fig:upsampled} shows some upsamled results of our L2G-AE.

\begin{figure}
    \centering
    \includegraphics[height=7cm,width=7cm]{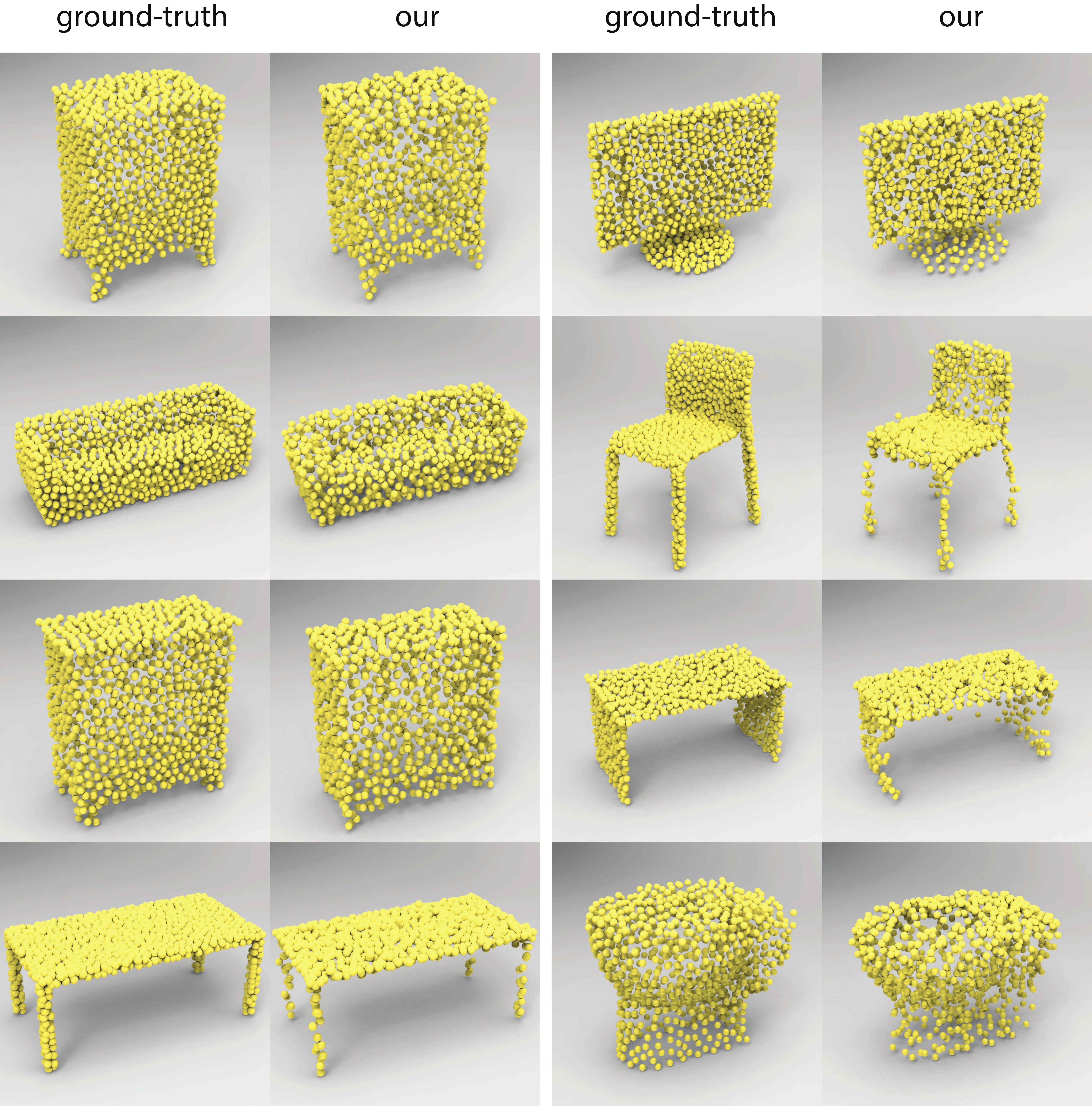}
    \caption{Some reconstructed examples of L2G-AE.}
    \label{fig:reconstruct}
\end{figure}
\begin{figure}
    \centering
    \includegraphics[height=10cm,width=7cm]{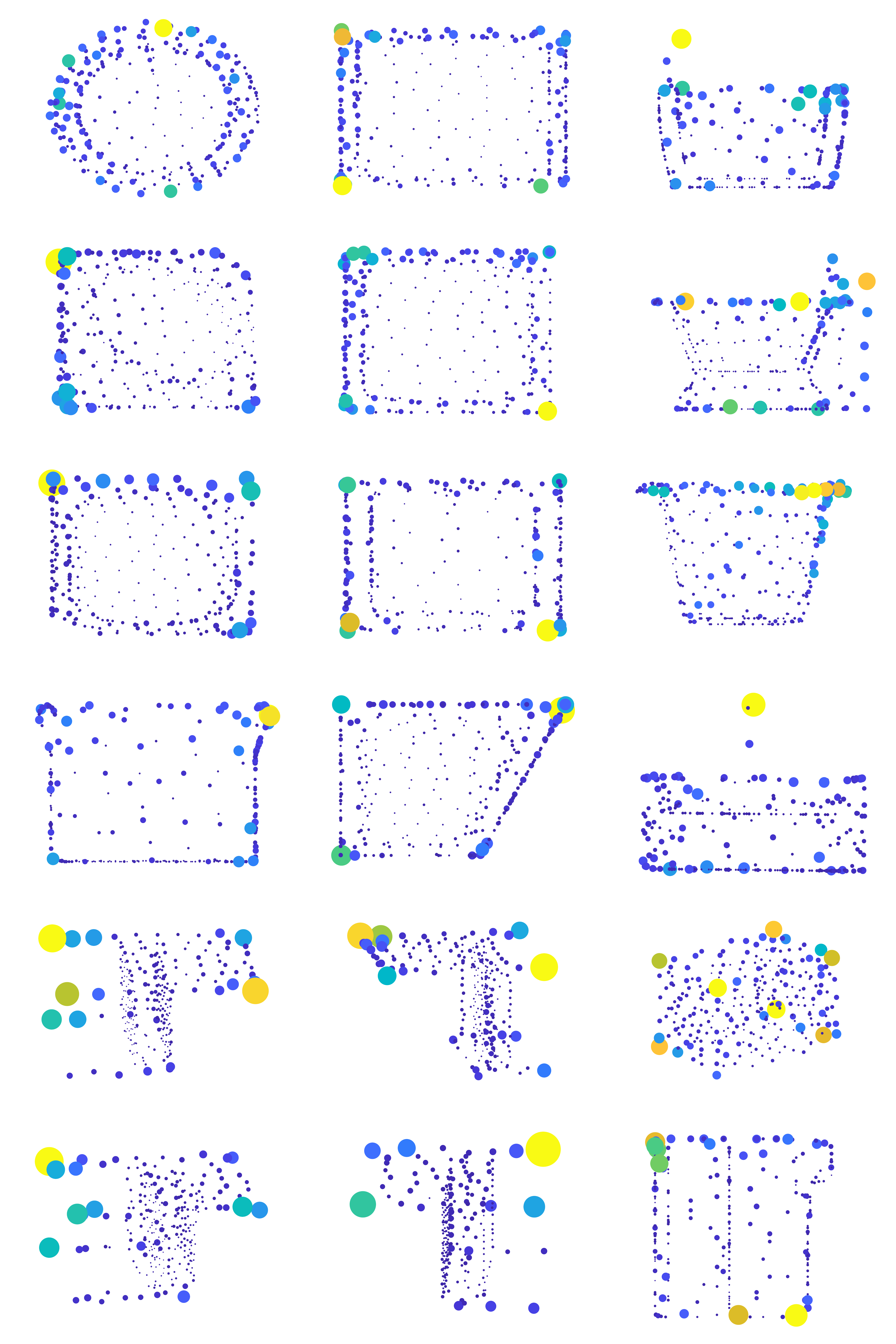}
    \caption{Some examples of the attention in the region level, where each subfigure represents a 3D object.}
    \label{fig:attetnion}
\end{figure}
\subsection{Visualization}
In this section,we will show some important visualization results of L2G-AE.
Firstly, some reconstructed point clouds by L2G-AE are listed with the ground-truths as shown in Figure \ref{fig:reconstruct}.
From the results, the reconstructed point clouds of L2G-AE are consistent with the ground-truths.

Then, some visualizations of the attention map inside self-attention modules are engaged to show the effect of attentions in the hierarchical feature abstraction.
There are three self-attention modules in the encoder, and we first visualize the attention map inside the local region level.
For  intuitively understanding, we directly attach the attention values to the centroids of local regions and then show these centroids.
By summing attention map by column in the region level, the attention value of each centroid is caculated. For example, a $256 \times 256$ attention map is translated to a 256-dimension attention vector, when the number of sampled centroids is 256.
Then, both the size and the color of centroids are associated with the attention values.
Therefore, the centroids with lighter colors and larger sizes indicate larger attention values.
As depicted in Figure \ref{fig:attetnion}, we show some examples of the region level attention.
Figure \ref{fig:attetnion} shows that the self-attention in the region level tends to on special local regions at conspicuous locations such as edges, corners or protruding parts.

\begin{figure}
  \centering
  \includegraphics[width=7cm,height=6cm]{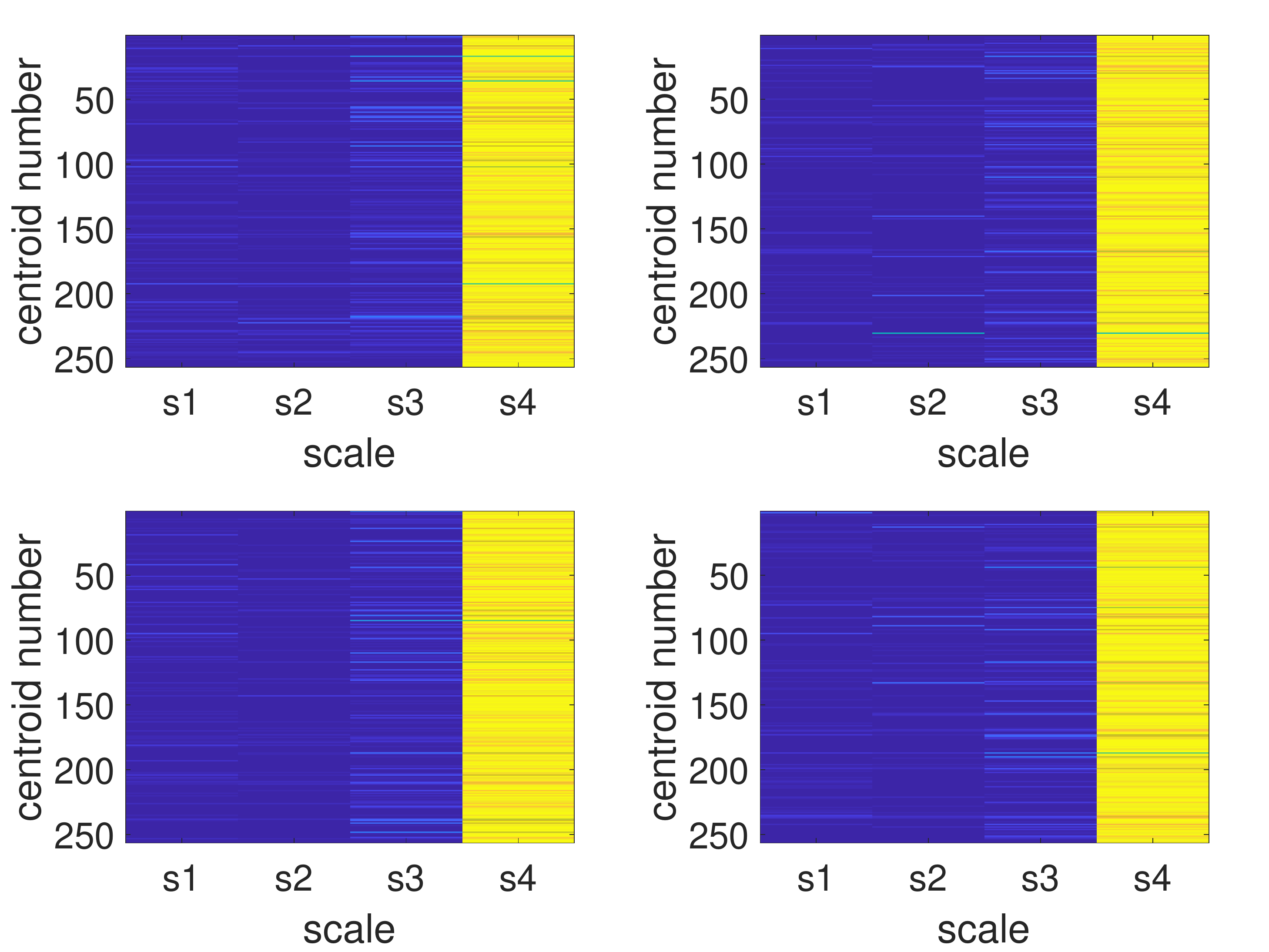}
  \caption{Some examples of the attention in the scale level. The abscissa represents the 4 scales $[s_1,s_2,s_3,s_4]$ around each centroid in a point cloud and the ordinate indicates the index of 256 centroids, where each subfigure represents a 3D object.}\label{fig:sl_attention}
\end{figure}
\begin{figure}
\centering
  \includegraphics[width=7cm,height=6cm]{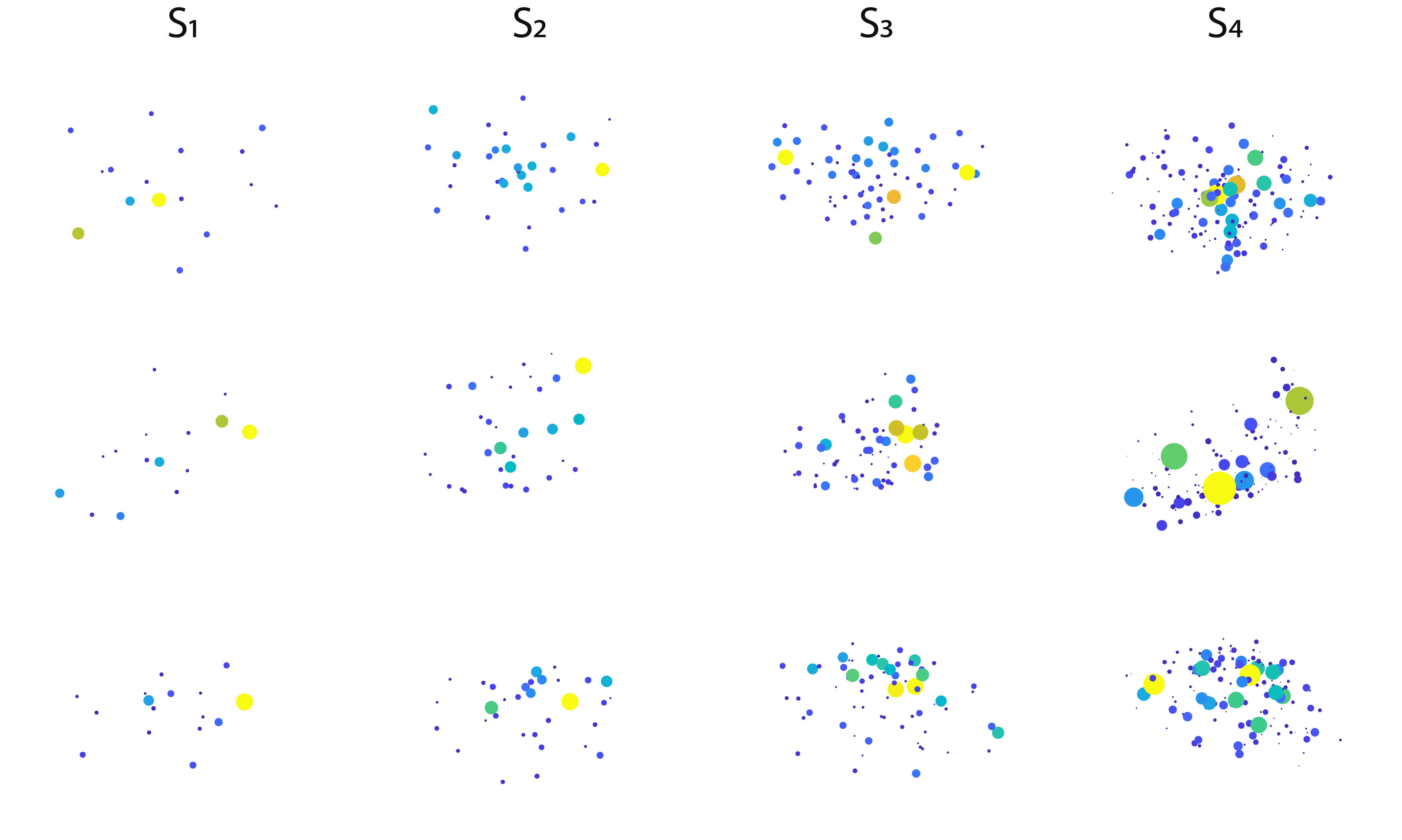}
  \caption{Some examples of the attention in the point level, where the four subfigures in each row represent the four scales of a local region.}\label{fig:pl_attention}
\end{figure}
Similarly,  we also show some examples of the scale level attention in Figure \ref{fig:sl_attention} and the point level attention in Figure \ref{fig:pl_attention}.
In Figure \ref{fig:sl_attention}, each image shows the 4 scale attention values around 256 sampled centroids of a point cloud. And the color indicates the value of attention, where large attention value corresponds to a bright color such as yellow.
The results indicate that the network tends to focus on the $4^{th}$ scale which contains more information of local structures.
In Figure \ref{fig:pl_attention}, each row represents the 4 scale areas around a centroid.
In different scale areas, the network concern on different points inside the areas to capture the local patterns in the local region.
\section{Conclusions}
In this paper, we propose a novel local to global Auto-encoder framework for point cloud understanding in the shape classification, retrieval and point cloud upsampling tasks.
In the encoder, a self-attention mechanism is employed to explore the correlation among features in the same level.
In contrast, an interpolation layer and RNN decoding layer successfully reconstruct local scales and global point clouds hierarchically.
Experimental results show that our method achieves competitive performances with state-of-the-art methods.

\bibliographystyle{ACM-Reference-Format}
\bibliography{reference}
\end{document}